\documentclass[sigconf]{acmart}
\usepackage{array}
\usepackage{multirow}

\AtBeginDocument{%
  }

\setcopyright{acmlicensed}
\copyrightyear{2025}
\acmYear{2025}
\acmConference[ASPDAC '25]{30th Asia and South Pacific Design Automation Conference}{January 20--23, 2025}{Tokyo, Japan}
\acmBooktitle{30th Asia and South Pacific Design Automation Conference (ASPDAC '25), January 20--23, 2025, Tokyo, Japan}\acmDOI{10.1145/3658617.3697721}
\acmISBN{979-8-4007-0635-6/25/01}

\begin{document}

\title{MTLSO: A Multi-Task Learning Approach for Logic Synthesis Optimization}

\author{Faezeh Faez}
\email{faezeh.faez@huawei.com}
\affiliation{%
  \institution{Huawei Noah’s Ark Lab}
  \city{Toronto}
  \country{Canada}
}

\author{Raika Karimi}
\email{raika.karimi@huawei.com}
\affiliation{%
  \institution{Huawei Noah’s Ark Lab}
  \city{Toronto}
  \country{Canada}
}

\author{Yingxue Zhang}
\email{yingxue.zhang@huawei.com}
\affiliation{%
  \institution{Huawei Noah’s Ark Lab}
  \city{Toronto}
  \country{Canada}
}

\author{Xing Li}
\email{li.xing2@huawei.com}
\affiliation{%
  \institution{Huawei Noah’s Ark Lab}
  \city{Hong Kong}
  \country{China}
}

\author{Lei Chen}
\email{lc.leichen@huawei.com}
\affiliation{%
  \institution{Huawei Noah’s Ark Lab}
  \city{Hong Kong}
  \country{China}
}

\author{Mingxuan Yuan}
\email{yuan.mingxuan@huawei.com}
\affiliation{%
  \institution{Huawei Noah’s Ark Lab}
  \city{Hong Kong}
  \country{China}
}

\author{Mahdi Biparva}
\email{mahdi.biparva@huawei.com}
\affiliation{%
  \institution{Huawei Noah’s Ark Lab}
  \city{Toronto}
  \country{Canada}
}

\renewcommand{\shortauthors}{Faez et al.}

\begin{abstract}
Electronic Design Automation (EDA) is essential for IC design and has recently benefited from AI-based techniques to improve efficiency. Logic synthesis, a key EDA stage, transforms high-level hardware descriptions into optimized netlists. Recent research has employed machine learning to predict Quality of Results (QoR) for pairs of And-Inverter Graphs (AIGs) and synthesis recipes. However, the severe scarcity of data due to a very limited number of available AIGs results in overfitting, significantly hindering performance. Additionally, the complexity and large number of nodes in AIGs make plain GNNs less effective for learning expressive graph-level representations. To tackle these challenges, we propose MTLSO - a Multi-Task Learning approach for Logic Synthesis Optimization. On one hand, it maximizes the use of limited data by training the model across different tasks. This includes introducing an auxiliary task of binary multi-label graph classification alongside the primary regression task, allowing the model to benefit from diverse supervision sources. On the other hand, we employ a hierarchical graph representation learning strategy to improve the model's capacity for learning expressive graph-level representations of large AIGs, surpassing traditional plain GNNs. Extensive experiments across multiple datasets and against state-of-the-art baselines demonstrate the superiority of our method, achieving an average performance gain of 8.22\% for delay and 5.95\% for area.
\end{abstract}



\maketitle

\section{Introduction}

Logic synthesis optimization (LSO) is a critical step in the electronic design automation (EDA) process, responsible for transforming a high-level hardware description language (HDL) representation of a circuit into an optimized netlist of Boolean logic gates. This transformation is essential for ensuring the final integrated circuit (IC) meets design criteria such as area and delay. Given the complexity and scale of modern ICs, which can contain billions of transistors, manual design is infeasible. EDA tools are essential for managing this complexity, but as designs grow more intricate, the traditional heuristic-based methods used in logic synthesis face challenges in achieving optimal results \cite{feng2022batch}. This highlights the need for more efficient and effective optimization methods. In this context, machine learning (ML) offers a promising avenue for enhancing LSO by providing faster and potentially more accurate predictions of QoR \cite{huang2021machine}.

Despite the potential of ML in EDA \cite{sohrabizadeh2023robust,yang2022versatile}, applying these techniques to LSO presents several challenges, particularly related to data scarcity and overfitting. The limited availability of large, labeled datasets in this domain hampers the ability to train robust ML models\cite{li2023verilog}, which in turn affects their generalization capability and prediction accuracy. This limitation impedes the practical deployment of ML models in production environments where they need to deliver consistent performance across diverse design scenarios.

To tackle overfitting, multi-task learning (MTL) has demonstrated significant promise in various domains such as image and text \cite{crawshaw2020multi}. MTL improves model generalization and robustness by leveraging shared representations across related tasks. By jointly learning multiple tasks, MTL mitigates data scarcity effects, benefiting from additional sources of supervision and enhancing performance across numerous applications.

To address the aforementioned issues, we propose \textbf{M}ulti-\textbf{T}ask Learning for \textbf{L}ogic \textbf{S}ynthesis \textbf{O}ptimization (MTLSO), an end-to-end, multi-task learning-based approach for logic synthesis. MTLSO introduces a novel task of binary multi-label graph classification alongside the primary task of QoR prediction, thereby optimizing the utilization of valuable yet limited data. We devised a novel approach to generate the required labels for this new task using existing data, eliminating the need for additional data. Furthermore, to enhance the efficiency of graph representation learning for large AIGs, we designed a hierarchical graph representation learning strategy. This strategy integrates GNNs with graph downsampling across multiple layers to facilitate learning more expressive graph-level representations based on multiple levels of abstraction. In summary, we present the following contributions in this work:
\begin{itemize}
    \item We design a novel task of binary multi-label graph classification alongside the conventional regression task. This multi-task training approach enables the model to learn from multiple sources of supervision, thereby improving its ability to predict the QoR for given pairs of AIGs and synthesis recipes.
    \item We introduce a label construction process that generates the necessary ground truth classification labels from existing data for training the graph classifier, eliminating the need for additional data sources.
    \item To enhance the expressiveness of the learned graph-level representations of large AIGs, we adopt a hierarchical strategy in our graph encoder, featuring multiple layers of successive graph encoding and downsampling.
    \item We conduct extensive experiments on several datasets and compare against state-of-the-art approaches to demonstrate the effectiveness of our proposed method, achieving an average gain of 8.22\% for delay and 5.95\% for area.
\end{itemize}
\section{Related Work}
In this section, we provide an overview of significant advancements related to our current work, organized into two subsections. First, we review the most notable machine learning-based approaches proposed to solve the LSO problem. Next, we delve into the accomplishments of multi-task learning across various domains, presenting a compelling rationale for adopting such a strategy to effectively address the LSO problem.
\subsection{Logic Synthesis Optimization}

Given the extreme complexity of VLSI chip designs, there has been a recent trend toward employing machine learning techniques to expedite design closure \cite{hamolia2021survey}. In this context, \citeauthor{yu2023machine} \cite{yu2023machine} examines the potential of machine learning models to enhance efficiency across various stages of EDA, including LSO. \textcolor{black}{FlowTune \cite{neto2022flowtune} leverages a domain-specific, multi-stage multi-armed bandit approach to explore and optimize synthesis toolflows.}
There are studies such as \cite{hosny2020drills} that compute representations of netlists by employing traditional techniques to extract hand-crafted features from AIGs. In contrast, other approaches utilize Graph Neural Networks (GNNs) as more advanced graph representation learners. These methodologies typically begin by employing a simple plain GNN to encode the input AIG. They subsequently employ various techniques to learn the representations of synthesis recipes. Prediction of QoR values for pairs of AIGs and recipes is then conducted by leveraging both the computed graph representations and the recipe representations. For instance, \citeauthor{chowdhury2021openabc} \cite{chowdhury2021openabc} learn representations of synthesis recipes by passing them through a set of 1D convolution layers. They then concatenate these representations with those learned for circuits by GNNs to predict QoR values. LOSTIN \cite{wu2022lostin} and GNN-H \cite{wu2022ai} use LSTM to learn representations of synthesis recipes, capturing the relative ordering of logic transformations within them. \citeauthor{yang2022prediction} \cite{yang2022prediction} adopt a similar strategy but replace the LSTM with the self-attention mechanism of the Transformer to learn representations of the recipes. None of these methods account for the varying importance of different graph nodes in learning the final graph-level representation of each AIG. They treat all graph nodes with equal importance, resulting in inefficiencies and reduced expressiveness of the final representation, particularly for very large AIGs, which are common in the LSO problem. Additionally, they have not addressed the critical issue of overfitting caused by severe data scarcity, which is a significant challenge in ML-based LSO solvers.
\subsection{Multi-task Learning}
Multi-task learning is a machine learning paradigm designed to enhance model generalization by utilizing shared data across multiple related tasks. This approach is particularly advantageous in scenarios characterized by data scarcity. Multi-task learning has demonstrated success in various fields, including natural language processing \cite{lim2020semi}, computer vision \cite{ding2023mitigating}, and speech recognition \cite{hu2023gradient}, among others. However, despite the significant challenges posed by data scarcity in LSO, which impede effective model training and underscore the need for MTL, it remains underutilized in this domain.

\section{Methodology}
\begin{figure*}[htbp]
  \centering
  \includegraphics[width=\linewidth]{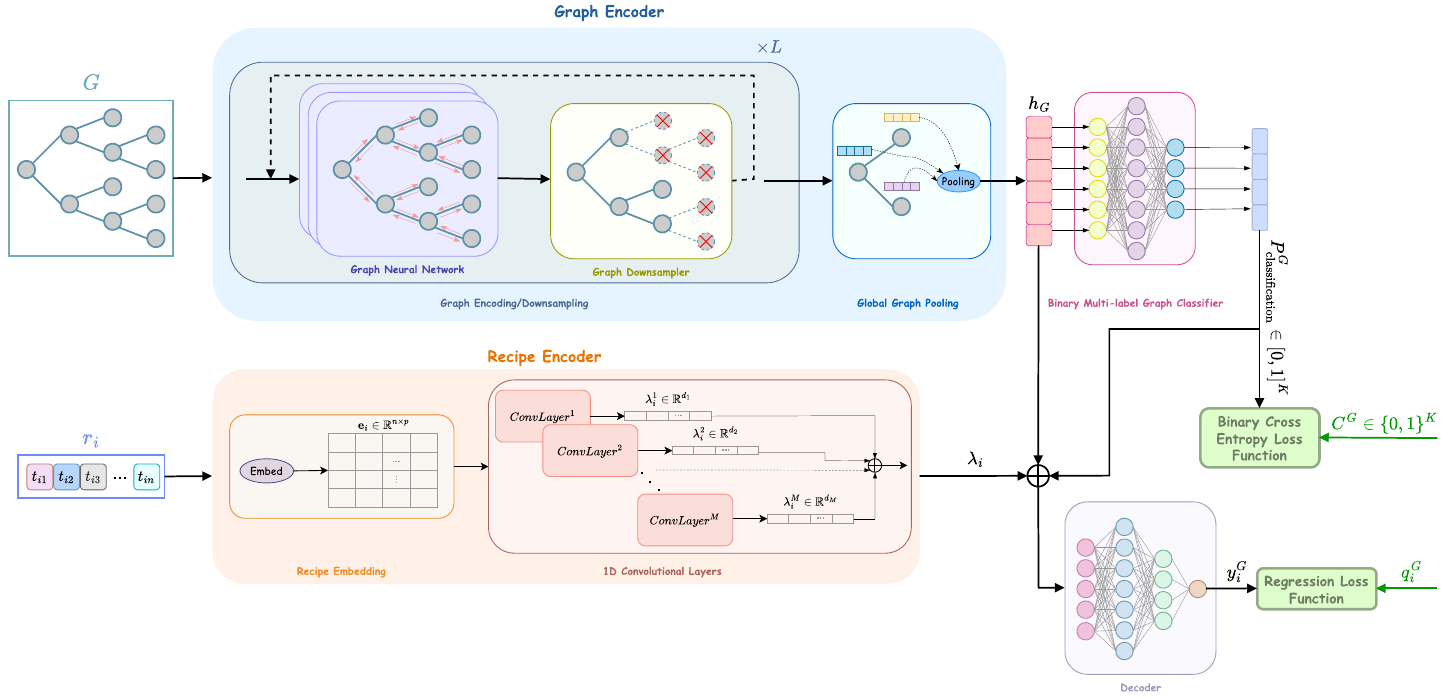}
  \Description{Overview of our MTLSO model which consists of Graph Encoder, Recipe Encoder, Binary Multi-label Graph Classifier, and Decoder. The model jointly optimizes graph classification and regression tasks.}
  \vspace{-20pt}
  \caption{Overview of our MTLSO model consisting of four main components: Graph Encoder, Recipe Encoder, Binary Multi-label Graph Classifier, and Decoder. The model parameters are jointly learned through multi-task learning, optimizing both graph classification and regression tasks, which enables the model to benefit from shared representations and inter-task dependencies.}
  \label{fig:overview}
\end{figure*}
In this section, we first define the problem formulation. Next, we introduce the main components of our MTLSO approach. Subsequently, we delve into a detailed explanation of our multi-task learning strategy, including its tasks, components, and objectives. We visualize an overview of our approach in Figure \ref{fig:overview}.

\subsection{Problem Formulation}
\textcolor{black}{Given an And-Inverter Graph (AIG) \( G \in \mathcal{G} \) and a set of \( K \) synthesis recipes \( \{r_i\}_{i=1}^{K} \), the goal of LSO is to predict the QoR value for each pair \( (G, r_i) \), which can then be further utilized to determine the best synthesis recipe for the graph \( G \). Formally, the aim is to learn the mapping function:}

\begin{equation}
    \textcolor{black}{f: \mathcal{G} \times \mathcal{R} \rightarrow \mathbb{R}}
\end{equation}

\textcolor{black}{where \( \mathcal{G} \) represents the set of AIGs and \( \mathcal{R} \) represents the set of all available synthesis recipes.}

\subsection{Graph Encoder}

Given a graph \( G \) with its node feature matrix \( X \), the goal of the graph encoder is to learn a graph-level representation \( h_G \) for the entire graph. This is typically accomplished by utilizing a Graph Neural Network (GNN), which initially learns a node representation matrix \( H \). These node representations are then aggregated, often using simple pooling techniques such as mean or max pooling, to derive the graph-level representation. To address the logic synthesis optimization problem, we adopt a hierarchical approach for learning graph representations. This is motivated by the inherent complexity of AIGs, where conventional plain GNNs may encounter limitations. In the following sections, we elaborate on our Hierarchical Graph Representation Learning (HGRL) strategy, providing both its motivation and workflow.

\textbf{Hierarchical Graph Representation Learning.} For solving the logic synthesis optimization problem, learning high-quality representations of AIGs is essential. These graphs vary in size, with most of them being large, having thousands or even tens of thousands of nodes. Furthermore, the importance of nodes in AIGs is not necessarily the same; some may play more critical roles in determining the final QoR value than others. As an example, some nodes may be redundant, meaning that removing them would not change the final logical function. Due to these reasons, solely using typical GNNs that conduct message passing among all the graph nodes in a flat manner may be less efficient. To address this challenge, we propose utilizing a hierarchical graph representation learning (HGRL) approach, which computes graph-level representations of AIGs at multiple levels of abstraction.

The HGRL consists of $L$ layers stacked sequentially. At each layer $l$, the graph $G^l$ is processed by a GNN, utilizing its node feature matrix $X^l \in \mathbb{R}^{N^l \times {F^l}}$ and adjacency matrix $A^l \in \{0, 1\}^{N^l \times N^l}$. Here, $N^l$ represents the number of nodes in $G^l$ and $F^l$ denotes the dimensionality of the node features at layer $l$. The initial AIG is represented as $G^0$, with $X^0$ and $A^0$ denoting its node feature matrix and adjacency matrix, respectively. The GNN at the $l$-th layer computes node representations through message passing among the $N^l$ nodes:
\begin{equation}
H^{l+1} = \text{GNN}(X^l, A^l) \label{eq:gnn_layer}
\end{equation}

where $H^{l+1} \in \mathbb{R}^{N^l \times F^{l+1}}$ is the representation matrix learned by the GNN for the nodes of $G^l$, and $F^{l+1}$ is the size of each learned node representation. After computing the node representations, we need to identify the top $N^{l+1} = \lceil \alpha N^l\rceil$ most important nodes, where $\alpha$ is a ratio hyperparameter. This choice is made by a graph downsampling module, which computes a score for each node based on the projection of the learned node representations \( H^{l+1} \) and a learnable vector. Nodes with lower scores are then removed from the graph \( G^l \). Thus, the pruned node representation matrix and the pruned adjacency matrix can be computed as follows:

\begin{equation}
    A^{l+1}, X^{l+1} = \text{\textsc{GraphDownsample}}(A^{l}, H^{l+1})
\end{equation}

where $A^{l+1} \in \{0, 1\}^{N^{l+1} \times N^{l+1}}$ is the pruned adjacency matrix at the end of the $l$-th layer, and $X^{l+1} \in \mathbb{R}^{N^{l+1} \times F^{l+1}}$ is the representation matrix for the $N^{l+1}$ remaining nodes. These two matrices are then given as inputs to the next encoding layer. After passing $A^0$ and $X^0$ through $L$ consecutive graph encoding blocks, where the output of each block serves as the input for the subsequent block, the final outputs $A^L \in \{0, 1\}^{N^L \times N^L}$ and $X^L \in \mathbb{R}^{N^L \times F^L}$ are computed. Here, $N^L$ represents the number of nodes remaining after all pruning iterations, while $F^L$ signifies the size of the final node representations. The final representation matrix $X^L$ is subsequently fed into a global pooling module to compute the graph representation. This process aggregates the representations learned for the most critical nodes after multiple steps of subsequent graph neural network operations and graph downsampling: 
\begin{equation}
    h_G = \text{\textsc{GraphPool}}(X^L)
\end{equation}
where $h_G \in \mathbb{R}^F$ denotes the graph-level representation of the input AIG, with $F$ representing its dimensionality.
\subsection{Recipe Encoder}
A synthesis recipe $r_i$ consists of a sequence of $n$ transformations, with each transformation falling into one of $m$ distinct categories. The sequence is represented as:

\begin{multline}
r_i = [t_{i1}, t_{i2}, \ldots, t_{in}], \\
\text{where } t_{ij} \in \{\mathcal{C}_1, \mathcal{C}_2, \ldots, \mathcal{C}_m\} \ \text{for } j = 1, 2, \ldots, n
\end{multline}

In this paper, we adopted the same set of transformations as described in \cite{chowdhury2021openabc}, where $m=7$ and the set $\{C_1, C_2, \ldots, C_7\}$ corresponds to \textbf{Balance (b)}, \textbf{Rewrite (rw, rw -z)}, and \textbf{Refactor (rf, rf -z)}.\par

To optimize logic synthesis by predicting the QoR for a given pair of AIG and recipe, it is essential to learn a meaningful representation of the recipe. In this subsection, we elucidate the steps to achieve this.

\subsubsection{Embedding}

The process of representation learning for a recipe $r_i$ commences with the conversion of its categorical transformations, each denoted as $t_{ij}$, into dense vector representations, which are low-dimensional continuous vectors. This conversion facilitates the model in identifying patterns and dependencies within the data:
\begin{equation}
    \mathbf{e}_i = \textsc{RecipeEmbed}(r_i)
\end{equation}
Here, $\mathbf{e}_i \in \mathbb{R}^{n \times p}$ denotes the resulting embedding, where $p$ signifies the dimensionality of the learned embedding for each transformation $t_{ij}$.
\subsubsection{Convolutional Layers}
Following the computation of the embedding $\mathbf{e}_i$ for the recipe $r_i$, a series of $M$ one-dimensional convolutional layers, each configured with a distinct kernel size, are employed. These layers serve to discern intricate relationships between adjacent transformations and extract features at multiple scales. The $m$-th convolutional layer is denoted as:
\begin{equation}
    \lambda^{m}_i = \textsc{ConvLayer}^{m}(\mathbf{e}_i)
\end{equation}
Here, $\lambda^{m}_i\in \mathbb{R}^{d_m}$ represents the $m$-th feature map learned for the recipe $r_i$, with a dimensionality of $d_m$. The final representation of recipe $r_i$ is computed by concatenating all $M$ feature maps associated with this recipe:
\begin{equation}
    \lambda_i = \text{Concat}(\lambda_i^1, \lambda_i^2, \ldots, \lambda_i^M)
\end{equation}
where $\lambda_i$ signifies the final representation learned for $r_i$.

\subsection{Multi-Task Learning} 
Addressing the LSO problem using AI-driven approaches is significantly challenged by data scarcity, as most datasets only contain a limited number of graphs. This limitation can result in model overfitting. To address this issue, we propose MTLSO, a multi-task learning approach that, in addition to the main task of QoR value regression, introduces a new task: binary multi-label graph classification. This supplementary task not only aids in training the model's parameters by providing additional signals during the training step but also signifies the relevance of a recipe to an AIG during inference, further assisting the model in predicting the QoR value. Further details on each task are discussed below.

\subsubsection{Binary Multi-label Graph Classification}

We have designed a novel task, binary multi-label graph classification, to address the LSO problem more effectively and to make the most usage out of the valuable yet limited training data. This task involves taking an AIG as input and determining whether each of the $K$ recipes is performing well on this AIG or not. In the following, we elaborate on the process of generating labels to be used for training the model and then formulate the task.

\textbf{Label Construction Process.} \textcolor{black}{For a given AIG $G$, the dataset provides QoR values for all $K$ associated recipes, denoted by $\{q^G_i\}_{i=1}^K$. The label construction process starts with selecting $\lceil \rho K \rceil$ recipes with the lowest QoR values (i.e., the best-performing recipes), where $\rho$ is a hyperparameter determining the ratio of top-performing recipes to be selected. Subsequently, the labels for these selected recipes are set to 1, while the labels for the remaining recipes are set to 0:}

\textcolor{black}{\[ c^G_i = \begin{cases} 
1 & \text{if $r_i$ is among the top-performing recipes of $G$} \\ 
0 & \text{otherwise} 
\end{cases} \]}

\textbf{Task Formulation.} \textcolor{black}{The graph classifier takes as input $h_G$, the graph-level representation learned for AIG $G$ by the graph encoder module. It then outputs a probability for each of the $K$ recipes, specifying whether each recipe is among the best for that AIG. Hence, the binary multi-label graph classification task is formulated as:}
\begin{equation}
\textcolor{black}{P^G_{\text{classification}} = \textsc{GraphClassify}(h_G)}
\end{equation}
\textcolor{black}{Here, $P^G_{\text{classification}} \in [0, 1]^K$ represents the predicted probabilities associated with the graph $G$, and the graph classifier is implemented as a 2-layer MLP with ReLU nonlinearity between the layers. The classification loss is computed as follows:}
\begin{equation}
\textcolor{black}{\mathcal{L}_{\text{classification}} = \textsc{BinaryCrossEntropy}(P^G_{\text{classification}}, C^G)}
\end{equation}
\textcolor{black}{where $C^G \in \{0, 1\}^K$ represents the true class labels for AIG $G$.}

\subsubsection{QoR Value Regression}

\textcolor{black}{In our MTLSO approach, the second and primary task entails predicting the QoR value for a given pair $(G, r_i)$. This prediction incorporates both the graph representation and the recipe representation, as well as the relative importance of the current recipe to the current graph, compared to other recipes, which is learned by the graph classifier. Consequently, the final QoR value for the pair $(G, r_i)$, denoted as $y^G_i$, is computed as follows:}

\begin{equation}
\textcolor{black}{y^G_i = \textsc{Decoder}\big(\text{Concat}(h_G, \lambda_i, P_{\text{classification}}^G)\big)}
\end{equation}
\textcolor{black}{The $\textsc{Decoder}$ in the formula above is implemented as a 3-layer MLP with ReLU nonlinearity. For computing the loss in this regression task, we employ the following function:}

\begin{equation}
\textcolor{black}{\mathcal{L}_{\text{regression}} = \textsc{RegressionLoss}(y^G_i, q^G_i)}
\end{equation}

\textcolor{black}{The detail regarding the choice of $\textsc{RegressionLoss}$ is provided in Section \ref{sec:exp}. Finally, the total loss function for training the model combines the classification loss and the regression loss:}

\begin{equation}
\textcolor{black}{\mathcal{L} = \mathcal{L}_{\text{classification}} + \gamma \mathcal{L}_{\text{regression}}}
\end{equation}
\textcolor{black}{where $\gamma$ is a weight hyperparameter to balance $\mathcal{L}_{\text{classification}}$ and $\mathcal{L}_{\text{regression}}$.}

\section{Experiments}\label{sec:exp}
\label{experiments}
\textcolor{black}{In this section, we first elaborate on the datasets used for the evaluation process. Then, we outline the state-of-the-art baselines with which we compare our method. Next, the evaluation metrics are explained, followed by a description of the experimental setup. Subsequently, we discuss the results of our proposed approach, as well as those of the competitor methods. Finally, we review the results of our ablation study on the role of each model component.}

\subsection{Datasets}
\textcolor{black}{To assess the performance of our proposed methodology, we conducted experiments using three datasets: OpenABC-D \cite{chowdhury2021openabc}, EPFL \cite{amaru2015epfl}, and a proprietary dataset, which we refer to as Commercial Dataset (CD) in this paper. Detailed statistics for these datasets are provided in Table \ref{tab:dataset}.}

\begin{table}
\begin{center}
\caption{\label{tab:dataset} \textbf{Statistics of datasets used in the experiments.}}
\textcolor{black}{
\resizebox{0.8\columnwidth}{!}{%
\begin{tabular}{|lm{0.08\textwidth}m{0.08\textwidth}m{0.08\textwidth}m{0.08\textwidth}|}
\hline
Dataset      & Min. \#Nodes                                   & Max. \#Nodes &     Avg. \#Nodes      &  \#Graphs      \\ \hline \hline
OpenABC-D          & 597 & 139719 & 36959.92 & 26      \\ \hline
EPFL      & 207 & 57503 & 15833.40 & 15     \\ \hline
CD         & 77 & 55332 & 21746.99 & 118      \\ \hline
\end{tabular}}}
\end{center}
\end{table}

\subsection{Baselines}
\textcolor{black}{We compare our approach with several well-known state-of-the-art methods, which are explained below.}
\begin{itemize}
    \item \textcolor{black}{\textbf{\citeauthor{chowdhury2021openabc} \cite{chowdhury2021openabc}}: They use GCN \cite{kipf2016semi} for encoding AIGs and a set of 1D convolutional layers to learn representations of recipes. These two representations are then concatenated to predict the QoR.} 
    \item \textcolor{black}{\textbf{LOSTIN \cite{wu2022lostin}:} This model employs GIN \cite{xu2018powerful} to compute representations of circuits and uses an LSTM for learning representations of synthesis recipes. The concatenation of these two representations is utilized for downstream QoR prediction.}
    \item \textcolor{black}{\textbf{GNN-H \cite{wu2022ai}:} It adopts a similar strategy for predicting QoR values as LOSTIN \cite{wu2022lostin}, except that it utilizes PNA \cite{corso2020principal} for circuit representation learning.}
    \item \textcolor{black}{\textbf{\citeauthor{yang2022prediction} \cite{yang2022prediction}:} They replace LSTM in LOSTIN \cite{wu2022lostin} and GNN-H \cite{wu2022ai} by a Transformer encoder. Moreover, they utilize GraphSage\cite{hamilton2017inductive} as the GNN.} 
\end{itemize}
\subsection{Metrics}

\textcolor{black}{To evaluate the effectiveness of our proposed method compared to other competitor approaches, we present the results in terms of Mean Absolute Percentage Error (MAPE). MAPE quantifies the accuracy of predictions by calculating the average absolute percentage difference between the actual and predicted values. It is defined as:}
\textcolor{black}{
\[
\text{MAPE} = \frac{100}{n} \sum_{i=1}^{n} \left| \frac{y_i - \hat{y}_i}{y_i} \right|
\]}

\textcolor{black}{where \( y_i \) denotes the actual value, \( \hat{y}_i \) denotes the predicted value, and \( n \) is the total number of observations.}

\subsection{Implementation Details}\label{subsec:imp}

Our implementation is done using PyTorch \cite{paszke2019pytorch}. The Graph Encoder module employed in our primary experiment comprises two layers of Graph Encoding/Downsampling (i.e., $L = 2$), a configuration further examined through an ablation study. We adopt a 2-layer GCN \cite{kipf2016semi} as our Graph Neural Network, with a hidden layer size of 64. The dimensions of the learned node features in the first and second encoding layers, denoted as $F^1$ and $F^2$, respectively, are both set to 64.

For the Graph Downsampler, TopKPooling \cite{cangea2018towards} is utilized with a node retainment ratio of 0.5 (i.e., $\alpha = 0.5$) for the primary experiments, with additional values examined in the ablation study. A multi-readout strategy that integrates both mean and max aggregations is adopted for \textsc{GraphPool}, resulting in a final graph-level representation dimension of 128 (i.e., $F = 128$).

The recipe encoder's embedding dimensionality is set to 60 (i.e., $p = 60$), and we utilize a series of four one-dimensional convolutional layers (i.e., $M = 4$). For label construction, the parameter $\rho$ is set to 0.5. Moreover, the Relative Squared Error (RSE) is used as the regression loss function. The hyperparameter $\gamma$ is set to 1, assigning equal weight to both graph classification and regression losses.

In our experimental setup, two-thirds of the graphs in each dataset are randomly allocated for model training, while the remaining one-third is reserved for testing. The evaluation results, presented in the following subsections (i.e., Subsections~\ref{sec:results} and~\ref{sec:ablation}), are reported on the test set.

\subsection{Results}\label{sec:results}
\begin{table*}[ht]
  \caption{Comparative Results in Terms of MAPE (Avg. $\pm$ Std.).}
  \label{main-results}
  \centering
  \renewcommand{\arraystretch}{1.2}  
  \setlength{\extrarowheight}{0pt}  
  \resizebox{\textwidth}{!}{%
    \begin{tabular}{|l|l|c|c|c|c|c|c|c|c|c|}
    \hline
    \multirow{2}{*}{Metric} & \multirow{2}{*}{Dataset} & \multicolumn{5}{c|}{Method} & \multicolumn{4}{c|}{Gain (\%)} \\
    \cline{3-11}
    & & \citeauthor{chowdhury2021openabc} \cite{chowdhury2021openabc} ($Y_1$) & LOSTIN \cite{wu2022lostin} ($Y_2$) & GNN-H \cite{wu2022ai} ($Y_3$) & \citeauthor{yang2022prediction} \cite{yang2022prediction} ($Y_4$) & MTLSO ($X$) & $\frac{Y_1 - X}{Y_1} \times 100$ & $\frac{Y_2 - X}{Y_2} \times 100$ & $\frac{Y_3 - X}{Y_3} \times 100$ & $\frac{Y_4 - X}{Y_4} \times 100$ \\
    \hline\hline
    \multirow{3}{*}{Delay} 
    & OpenABC-D & 23.66 $\pm$ 0.19 & 24.61 $\pm$ 0.03 & 24.31 $\pm$ 0.27 & 24.58 $\pm$ 0.13 & \textbf{22.93 $\pm$ 0.23} & 3.09 & 6.83 & 5.68 & 6.71 \\
    
    & EPFL & 4.18 $\pm$ 0.07 & 3.96 $\pm$ 0.02 & 3.96 $\pm$ 0.03 & 3.96 $\pm$ 0.02 & \textbf{3.94 $\pm$ 0.01} & 5.74 & 0.51 & 0.51 & 0.51 \\
    
    & CD & 15.88 $\pm$ 0.41 & 16.76 $\pm$ 0.13 & 16.80 $\pm$ 0.06 & 17.09 $\pm$ 0.08 & \textbf{13.75 $\pm$ 0.25} & 13.41 & 17.96 & 18.15 & 19.54 \\
    \hline
    \multirow{3}{*}{Area}  
    & OpenABC-D & 2.71 $\pm$ 0.02 & 2.35 $\pm$ 0.07 & \textbf{2.33 $\pm$ 0.06} & 3.77 $\pm$ 0.00 & 2.57 $\pm$ 0.03 & 5.17 & -9.36 & -10.30 & 31.83 \\
    
    & EPFL & 2.46 $\pm$ 0.03 & 2.30 $\pm$ 0.01 & 2.34 $\pm$ 0.02 & 2.39 $\pm$ 0.00 & \textbf{2.23 $\pm$ 0.04} & 9.35 & 3.04 & 4.70 & 6.69 \\
    
    & CD & 3.57 $\pm$ 0.10 & 3.46 $\pm$ 0.17 & 3.59 $\pm$ 0.12 & 3.81 $\pm$ 0.05 & \textbf{3.33 $\pm$ 0.08} & 6.72 & 3.76 & 7.24 & 12.60 \\
    \hline
    \end{tabular}%
  }
\end{table*}

\textcolor{black}{The main results, reported in Table \ref{main-results}, are averaged over multiple experimental runs. These findings demonstrate that our proposed MTLSO method outperforms the four baseline models in nearly all scenarios, with average gains of 8.22\% in delay and 5.95\% in area across all baselines and datasets. This underscores the critical advantage of employing a multi-task learning approach combined with a hierarchical graph encoder, rather than relying on simpler alternatives, for effectively addressing this problem. More specifically, the integration of an additional classification task contributed significantly to these gains by introducing increased supervision during training. This strategy effectively leverages the limited yet highly valuable labeled data, thereby reducing overfitting and enhancing the model’s ability to generalize to unseen test data, as shown in Table \ref{main-results}.}

\textcolor{black}{It is important to note that our main architecture shares similarities with the one proposed by \citeauthor{chowdhury2021openabc} \cite{chowdhury2021openabc}, particularly in our decision to use less complex techniques such as GCN \cite{kipf2016semi} as the core of our graph encoder and 1D convolution layers for encoding the recipes. This contrasts with the more advanced recipe encoders, like LSTM or Transformer, employed by LOSTIN \cite{wu2022lostin}, GNN-H \cite{wu2022ai}, and \citeauthor{yang2022prediction} \cite{yang2022prediction}, which consider the ordering of logic transformations within each recipe. We opted for simpler model components to demonstrate that multi-task learning powered by HGRL, even with these basic components, can outperform the baselines. This suggests that performance could be further enhanced by using more advanced GNNs like GIN, as utilized by LOSTIN \cite{wu2022lostin}, or by incorporating more sophisticated recipe encoders.}

\textcolor{black}{In the next subsection, we conduct an ablation study to analyze the individual contributions of the multi-task learning approach and the hierarchical graph encoder to the overall model performance.}

\subsection{Ablation Study}\label{sec:ablation}
\textcolor{black}{To assess the importance of each part of our model, we conducted an ablation study. In this study, we first modified our MTLSO method by replacing our Hierarchical Graph Representation Learning (HGRL) module with a simpler version, referred to as Plain Graph Representation Learning (PGRL). PGRL consists of a single GCN layer without any downsampling layer. Next, we retained the HGRL module from our MTLSO method but eliminated the graph classification task, training the model in a Single-task Learning (STL) mode. We present the results in Table \ref{ablation1}. As shown in the table, MTLSO achieves the best results, indicating that both the HGRL module and the multi-task learning strategy significantly contribute to the model's performance. When comparing the significance of multi-task learning and the HGRL module, the results demonstrate that the former is more critical than the latter. This is evidenced by the fact that PGRL (trained in a multi-task manner) outperforms the STL variant.}

\begin{table}
  \caption{\textcolor{black}{Ablation Study Results of Model Components in Terms of MAPE.}}
  \label{ablation1}
  \centering
  \resizebox{\columnwidth}{!}{%
    \begin{tabular}{@{}ccccccccc@{}}
    \toprule 
    & \multicolumn{3}{c}{Delay} & \multicolumn{3}{c}{Area} \\
    \cmidrule(lr){2-4} \cmidrule(lr){5-7}
    & \multicolumn{1}{c}{OpenABC-D} & \multicolumn{1}{c}{EPFL}& \multicolumn{1}{c}{CD} & \multicolumn{1}{c}{OpenABC-D} & \multicolumn{1}{c}{EPFL}& \multicolumn{1}{c}{CD} \\
    \midrule
    PGRL    & 23.49\% & 3.95\% & 16.48\% & 2.99\% & 2.24\% & 3.44\% \\
    STL    & 23.61\% & 4.51\% & 15.56\% & 2.69\% & 2.57\% & 3.45\% \\
    MTLSO    & \textbf{22.93\%} & \textbf{3.94\%} & \textbf{13.75\%} & \textbf{2.57\%} & \textbf{2.23\%} & \textbf{3.33\%}   \\
    \bottomrule
  \end{tabular}
  }
\end{table}

\begin{figure}[b]
  \centering
  \includegraphics[width=\columnwidth]{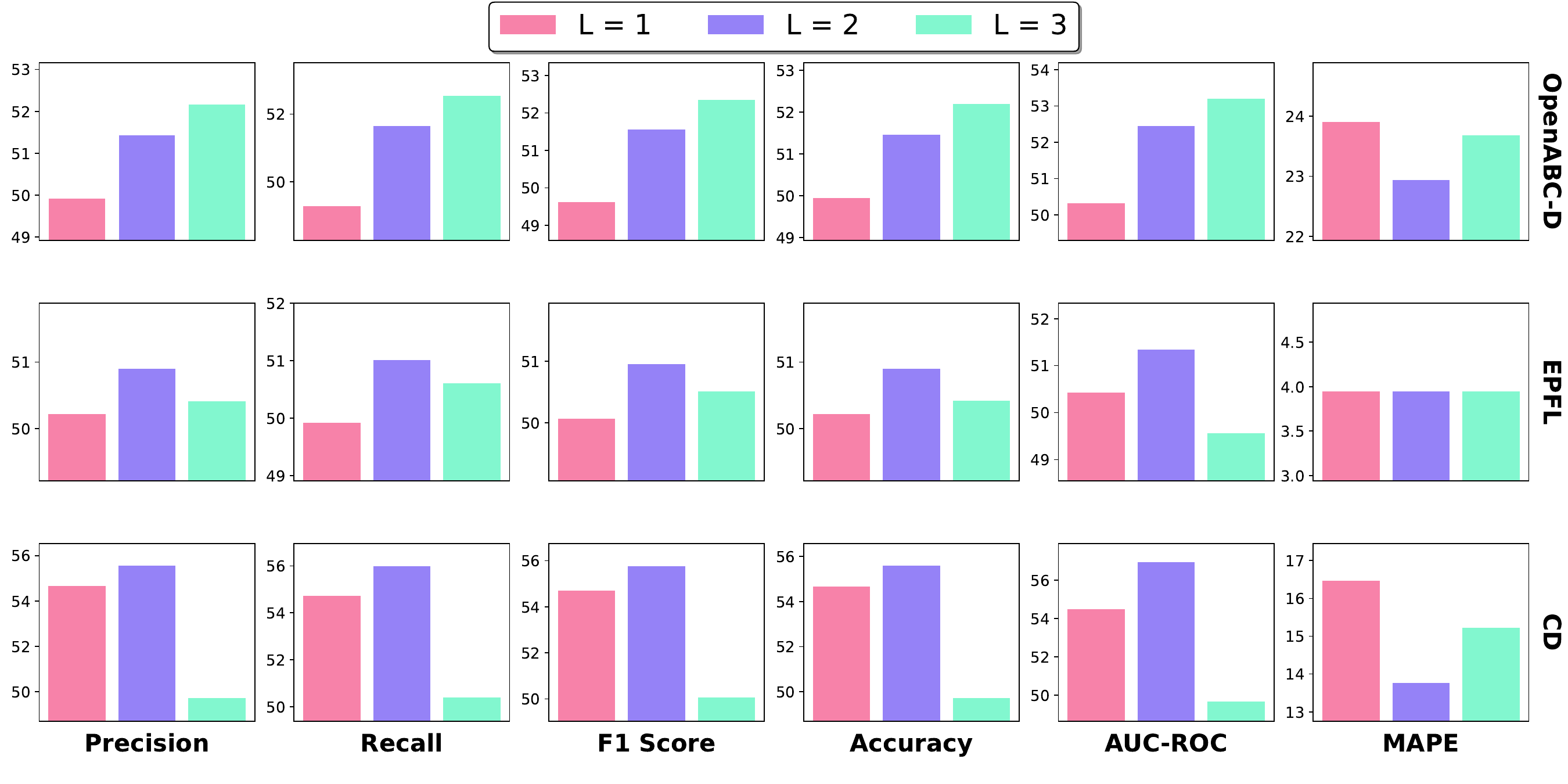}
  \Description{Ablation Study results showing the effect of varying the number of successive Graph Encoding/Downsampling Layers (L) on the performance metrics. Each row of subplots corresponds to one dataset, and each column corresponds to a specific evaluation metric.}
  \vspace{-20pt}
  \caption{\textcolor{black}{Ablation Study on the Number of Successive Graph Encoding/Downsampling Layers ($L$). The results are reported as percentages. Each row of subplots corresponds to one dataset, and each column corresponds to an evaluation metric.}}
  \label{fig:ablation}
\end{figure}

\textcolor{black}{We performed an additional ablation study on the number of Graph Encoding/Downsampling layers, denoted as \( L \). In addition to our main experiments with \( L=2 \), we investigated the cases where \( L \) was set to \( 1 \) and \( 3 \), keeping all other hyperparameters consistent with those reported in Subsection \ref{subsec:imp}. The results for delay minimization are presented in Figure \ref{fig:ablation}. We assessed performance using graph classification metrics and MAPE to understand how the hyperparameter \( L \) influences both the quality of the learned graph-level representation and the efficiency of the regression task. The results indicate that performance consistently improves with more than one encoding layer. This underscores the necessity of adopting a hierarchical strategy for effectively learning representations of large AIGs. Moreover, as the results suggest, the optimal \( L \) is 2 for EPFL and CD, and 3 for OpenABC-D based on the graph classification metrics. This implies that we can further improve the quality of graph representations by using the optimal value of the hyperparameter $L$ for each individual dataset.}

\begin{figure}[H]
  \centering
  \includegraphics[width=\columnwidth]{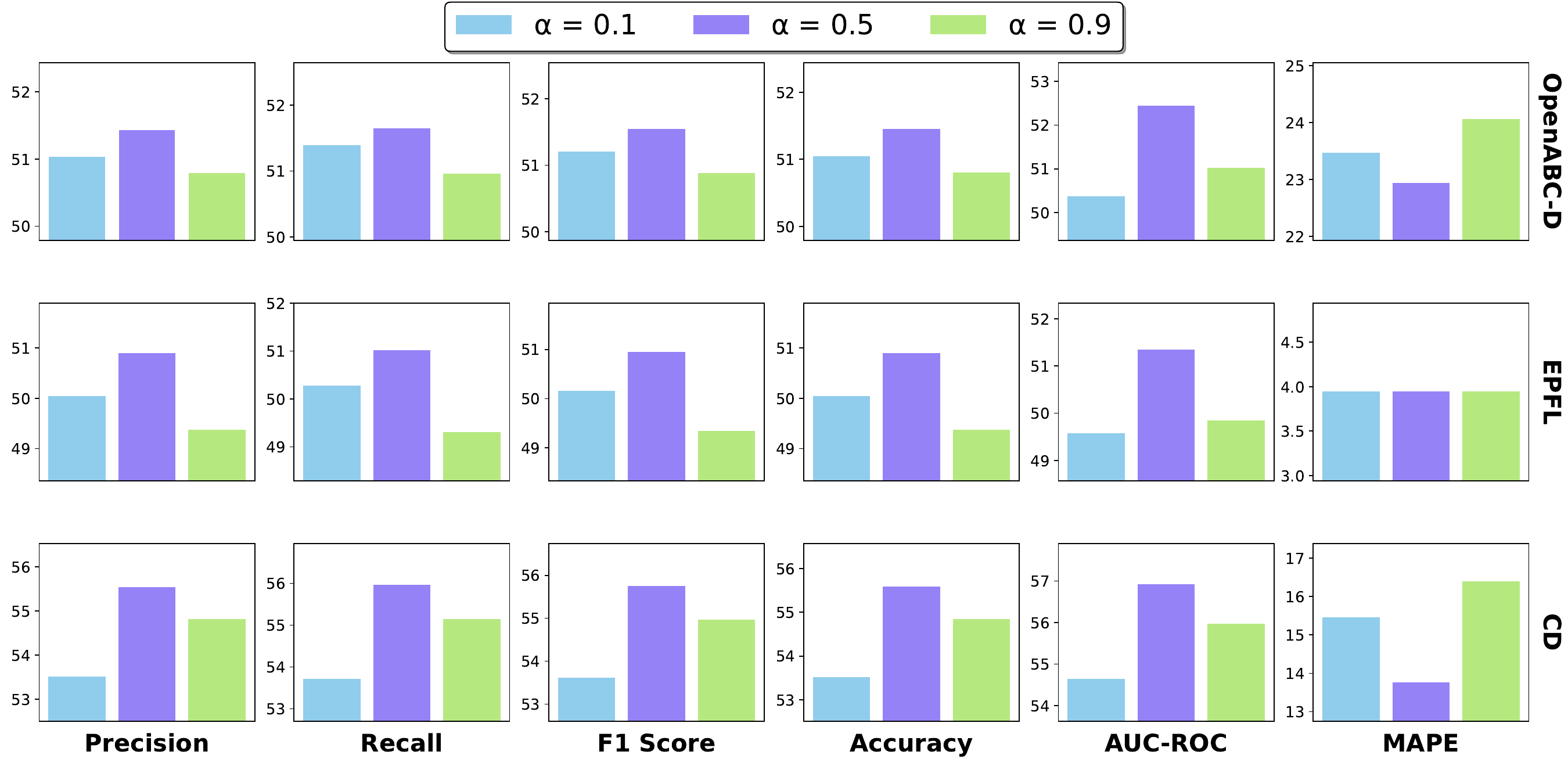}
  \Description{Ablation Study on the Node Retainment Ratio (alpha) of the Graph Downsampling Module. Each row of subplots corresponds to one dataset, and each column corresponds to a specific evaluation metric.}
  \vspace{-20pt}
  \caption{\textcolor{black}{Ablation Study on the Node Retainment Ratio ($\alpha$) of the Graph Downsampling Module. The results are reported as percentages. Each row of subplots corresponds to one dataset, and each column corresponds to an evaluation metric.}}
  \label{fig:ablation2}
\end{figure}

\textcolor{black}{Finally, an ablation study was conducted on the ratio hyperparameter $\alpha$, which determines the percentage of nodes retained by the graph downsampling module during each graph encoding block. The results, illustrated in charts within Figure \ref{fig:ablation2}, encompass three values of $\alpha$: specifically, 0.5 alongside the extremes of 0.1 and 0.9. All other hyperparameters were maintained as specified in Subsection~\ref{subsec:imp} (e.g., $L=2$). Across these experiments, optimal performance consistently favored $\alpha=0.5$, with both extreme values showing inferior results. This finding supports the importance of adopting such a hierarchical strategy for encoding AIGs, since a portion of nodes are less informative and should be filtered out (as evidenced by $\alpha=0.9$ performing worse than $\alpha=0.5$). Conversely, the underperformance of $\alpha=0.1$ highlights that some nodes have a direct influence on the quality of the final graph-level representation, and hence should not be pruned. This underscores the need to set an optimal value for this hyperparameter.}

\balance

\section{Conclusion}

\textcolor{black}{In this paper, we present a novel multi-task learning approach designed for LSO, focusing on mitigating overfitting caused by limited data availability, a critical issue of machine learning-based LSO solvers. Our approach involves jointly training a model for both multi-label graph classification and regression tasks, enhancing its ability to predict QoR values for pairs of AIGs and synthesis recipes. To further boost the effectiveness of our method, we employ a hierarchical approach with multiple layers of successive Graph Encoding/Downsampling to learn graph-level representations of AIGs, as due to the inherent complexity and large size of these graphs, simple plain GNNs often struggle to be efficient enough. Extensive experiments across multiple datasets, compared against four established baselines, demonstrate an average performance gain of 8.22\% and 5.95\% for delay and area minimization, respectively, underscoring the effectiveness of our proposed approach.}

\bibliographystyle{ACM-Reference-Format}
\bibliography{references.bib}


\begin{thebibliography}{24}


\ifx \showCODEN    \undefined \def \showCODEN     #1{\unskip}     \fi
\ifx \showDOI      \undefined \def \showDOI       #1{#1}\fi
\ifx \showISBNx    \undefined \def \showISBNx     #1{\unskip}     \fi
\ifx \showISBNxiii \undefined \def \showISBNxiii  #1{\unskip}     \fi
\ifx \showISSN     \undefined \def \showISSN      #1{\unskip}     \fi
\ifx \showLCCN     \undefined \def \showLCCN      #1{\unskip}     \fi
\ifx \shownote     \undefined \def \shownote      #1{#1}          \fi
\ifx \showarticletitle \undefined \def \showarticletitle #1{#1}   \fi
\ifx \showURL      \undefined \def \showURL       {\relax}        \fi
\providecommand\bibfield[2]{#2}
\providecommand\bibinfo[2]{#2}
\providecommand\natexlab[1]{#1}
\providecommand\showeprint[2][]{arXiv:#2}

\bibitem[Amar{\'u} et~al\mbox{.}(2015)]%
        {amaru2015epfl}
\bibfield{author}{\bibinfo{person}{Luca Amar{\'u}}, \bibinfo{person}{Pierre-Emmanuel Gaillardon}, {and} \bibinfo{person}{Giovanni De~Micheli}.} \bibinfo{year}{2015}\natexlab{}.
\newblock \showarticletitle{The EPFL combinational benchmark suite}. In \bibinfo{booktitle}{\emph{Proceedings of the 24th International Workshop on Logic \& Synthesis (IWLS)}}.
\newblock


\bibitem[Cangea et~al\mbox{.}(2018)]%
        {cangea2018towards}
\bibfield{author}{\bibinfo{person}{C{\u{a}}t{\u{a}}lina Cangea}, \bibinfo{person}{Petar Veli{\v{c}}kovi{\'c}}, \bibinfo{person}{Nikola Jovanovi{\'c}}, \bibinfo{person}{Thomas Kipf}, {and} \bibinfo{person}{Pietro Li{\`o}}.} \bibinfo{year}{2018}\natexlab{}.
\newblock \showarticletitle{Towards sparse hierarchical graph classifiers}.
\newblock \bibinfo{journal}{\emph{arXiv preprint arXiv:1811.01287}} (\bibinfo{year}{2018}).
\newblock


\bibitem[Chowdhury et~al\mbox{.}(2021)]%
        {chowdhury2021openabc}
\bibfield{author}{\bibinfo{person}{Animesh~Basak Chowdhury}, \bibinfo{person}{Benjamin Tan}, \bibinfo{person}{Ramesh Karri}, {and} \bibinfo{person}{Siddharth Garg}.} \bibinfo{year}{2021}\natexlab{}.
\newblock \showarticletitle{Openabc-d: A large-scale dataset for machine learning guided integrated circuit synthesis}.
\newblock \bibinfo{journal}{\emph{arXiv preprint arXiv:2110.11292}} (\bibinfo{year}{2021}).
\newblock


\bibitem[Corso et~al\mbox{.}(2020)]%
        {corso2020principal}
\bibfield{author}{\bibinfo{person}{Gabriele Corso}, \bibinfo{person}{Luca Cavalleri}, \bibinfo{person}{Dominique Beaini}, \bibinfo{person}{Pietro Li{\`o}}, {and} \bibinfo{person}{Petar Veli{\v{c}}kovi{\'c}}.} \bibinfo{year}{2020}\natexlab{}.
\newblock \showarticletitle{Principal neighbourhood aggregation for graph nets}.
\newblock \bibinfo{journal}{\emph{Advances in Neural Information Processing Systems}}  \bibinfo{volume}{33} (\bibinfo{year}{2020}), \bibinfo{pages}{13260--13271}.
\newblock


\bibitem[Crawshaw(2020)]%
        {crawshaw2020multi}
\bibfield{author}{\bibinfo{person}{Michael Crawshaw}.} \bibinfo{year}{2020}\natexlab{}.
\newblock \showarticletitle{Multi-task learning with deep neural networks: A survey}.
\newblock \bibinfo{journal}{\emph{arXiv preprint arXiv:2009.09796}} (\bibinfo{year}{2020}).
\newblock


\bibitem[Ding et~al\mbox{.}(2023)]%
        {ding2023mitigating}
\bibfield{author}{\bibinfo{person}{Chuntao Ding}, \bibinfo{person}{Zhichao Lu}, \bibinfo{person}{Shangguang Wang}, \bibinfo{person}{Ran Cheng}, {and} \bibinfo{person}{Vishnu~Naresh Boddeti}.} \bibinfo{year}{2023}\natexlab{}.
\newblock \showarticletitle{Mitigating task interference in multi-task learning via explicit task routing with non-learnable primitives}. In \bibinfo{booktitle}{\emph{Proceedings of the IEEE/CVF Conference on Computer Vision and Pattern Recognition}}. \bibinfo{pages}{7756--7765}.
\newblock


\bibitem[Feng et~al\mbox{.}(2022)]%
        {feng2022batch}
\bibfield{author}{\bibinfo{person}{Chang Feng}, \bibinfo{person}{Wenlong Lyu}, \bibinfo{person}{Zhitang Chen}, \bibinfo{person}{Junjie Ye}, \bibinfo{person}{Mingxuan Yuan}, {and} \bibinfo{person}{Jianye Hao}.} \bibinfo{year}{2022}\natexlab{}.
\newblock \showarticletitle{Batch sequential black-box optimization with embedding alignment cells for logic synthesis}. In \bibinfo{booktitle}{\emph{Proceedings of the 41st IEEE/ACM International Conference on Computer-Aided Design}}. \bibinfo{pages}{1--9}.
\newblock


\bibitem[Hamilton et~al\mbox{.}(2017)]%
        {hamilton2017inductive}
\bibfield{author}{\bibinfo{person}{Will Hamilton}, \bibinfo{person}{Zhitao Ying}, {and} \bibinfo{person}{Jure Leskovec}.} \bibinfo{year}{2017}\natexlab{}.
\newblock \showarticletitle{Inductive representation learning on large graphs}.
\newblock \bibinfo{journal}{\emph{Advances in neural information processing systems}}  \bibinfo{volume}{30} (\bibinfo{year}{2017}).
\newblock


\bibitem[Hamolia and Melnyk(2021)]%
        {hamolia2021survey}
\bibfield{author}{\bibinfo{person}{Vladyslav Hamolia} {and} \bibinfo{person}{Viktor Melnyk}.} \bibinfo{year}{2021}\natexlab{}.
\newblock \showarticletitle{A survey of machine learning methods and applications in electronic design automation}. In \bibinfo{booktitle}{\emph{2021 11th International conference on advanced computer information technologies (ACIT)}}. IEEE, \bibinfo{pages}{757--760}.
\newblock


\bibitem[Hosny et~al\mbox{.}(2020)]%
        {hosny2020drills}
\bibfield{author}{\bibinfo{person}{Abdelrahman Hosny}, \bibinfo{person}{Soheil Hashemi}, \bibinfo{person}{Mohamed Shalan}, {and} \bibinfo{person}{Sherief Reda}.} \bibinfo{year}{2020}\natexlab{}.
\newblock \showarticletitle{DRiLLS: Deep reinforcement learning for logic synthesis}. In \bibinfo{booktitle}{\emph{2020 25th Asia and South Pacific Design Automation Conference (ASP-DAC)}}. IEEE, \bibinfo{pages}{581--586}.
\newblock


\bibitem[Hu et~al\mbox{.}(2023)]%
        {hu2023gradient}
\bibfield{author}{\bibinfo{person}{Yuchen Hu}, \bibinfo{person}{Chen Chen}, \bibinfo{person}{Ruizhe Li}, \bibinfo{person}{Qiushi Zhu}, {and} \bibinfo{person}{Eng~Siong Chng}.} \bibinfo{year}{2023}\natexlab{}.
\newblock \showarticletitle{Gradient remedy for multi-task learning in end-to-end noise-robust speech recognition}. In \bibinfo{booktitle}{\emph{ICASSP 2023-2023 IEEE International Conference on Acoustics, Speech and Signal Processing (ICASSP)}}. IEEE, \bibinfo{pages}{1--5}.
\newblock


\bibitem[Huang et~al\mbox{.}(2021)]%
        {huang2021machine}
\bibfield{author}{\bibinfo{person}{Guyue Huang}, \bibinfo{person}{Jingbo Hu}, \bibinfo{person}{Yifan He}, \bibinfo{person}{Jialong Liu}, \bibinfo{person}{Mingyuan Ma}, \bibinfo{person}{Zhaoyang Shen}, \bibinfo{person}{Juejian Wu}, \bibinfo{person}{Yuanfan Xu}, \bibinfo{person}{Hengrui Zhang}, \bibinfo{person}{Kai Zhong}, {et~al\mbox{.}}} \bibinfo{year}{2021}\natexlab{}.
\newblock \showarticletitle{Machine learning for electronic design automation: A survey}.
\newblock \bibinfo{journal}{\emph{ACM Transactions on Design Automation of Electronic Systems (TODAES)}} \bibinfo{volume}{26}, \bibinfo{number}{5} (\bibinfo{year}{2021}), \bibinfo{pages}{1--46}.
\newblock


\bibitem[Kipf and Welling(2016)]%
        {kipf2016semi}
\bibfield{author}{\bibinfo{person}{Thomas~N Kipf} {and} \bibinfo{person}{Max Welling}.} \bibinfo{year}{2016}\natexlab{}.
\newblock \showarticletitle{Semi-supervised classification with graph convolutional networks}.
\newblock \bibinfo{journal}{\emph{arXiv preprint arXiv:1609.02907}} (\bibinfo{year}{2016}).
\newblock


\bibitem[Li et~al\mbox{.}(2023)]%
        {li2023verilog}
\bibfield{author}{\bibinfo{person}{Yingjie Li}, \bibinfo{person}{Mingju Liu}, \bibinfo{person}{Alan Mishchenko}, {and} \bibinfo{person}{Cunxi Yu}.} \bibinfo{year}{2023}\natexlab{}.
\newblock \showarticletitle{Verilog-to-PyG-A Framework for Graph Learning and Augmentation on RTL Designs}. In \bibinfo{booktitle}{\emph{2023 IEEE/ACM International Conference on Computer Aided Design (ICCAD)}}. IEEE, \bibinfo{pages}{1--4}.
\newblock


\bibitem[Lim et~al\mbox{.}(2020)]%
        {lim2020semi}
\bibfield{author}{\bibinfo{person}{KyungTae Lim}, \bibinfo{person}{Jay~Yoon Lee}, \bibinfo{person}{Jaime Carbonell}, {and} \bibinfo{person}{Thierry Poibeau}.} \bibinfo{year}{2020}\natexlab{}.
\newblock \showarticletitle{Semi-supervised learning on meta structure: Multi-task tagging and parsing in low-resource scenarios}. In \bibinfo{booktitle}{\emph{Proceedings of the AAAI Conference on Artificial Intelligence}}, Vol.~\bibinfo{volume}{34}. \bibinfo{pages}{8344--8351}.
\newblock


\bibitem[Neto et~al\mbox{.}(2022)]%
        {neto2022flowtune}
\bibfield{author}{\bibinfo{person}{Walter~Lau Neto}, \bibinfo{person}{Yingjie Li}, \bibinfo{person}{Pierre-Emmanuel Gaillardon}, {and} \bibinfo{person}{Cunxi Yu}.} \bibinfo{year}{2022}\natexlab{}.
\newblock \showarticletitle{FlowTune: End-to-end automatic logic optimization exploration via domain-specific multiarmed bandit}.
\newblock \bibinfo{journal}{\emph{IEEE Transactions on Computer-Aided Design of Integrated Circuits and Systems}} \bibinfo{volume}{42}, \bibinfo{number}{6} (\bibinfo{year}{2022}), \bibinfo{pages}{1912--1925}.
\newblock


\bibitem[Paszke et~al\mbox{.}(2019)]%
        {paszke2019pytorch}
\bibfield{author}{\bibinfo{person}{Adam Paszke}, \bibinfo{person}{Sam Gross}, \bibinfo{person}{Francisco Massa}, \bibinfo{person}{Adam Lerer}, \bibinfo{person}{James Bradbury}, \bibinfo{person}{Gregory Chanan}, \bibinfo{person}{Trevor Killeen}, \bibinfo{person}{Zeming Lin}, \bibinfo{person}{Natalia Gimelshein}, \bibinfo{person}{Luca Antiga}, {et~al\mbox{.}}} \bibinfo{year}{2019}\natexlab{}.
\newblock \showarticletitle{Pytorch: An imperative style, high-performance deep learning library}.
\newblock \bibinfo{journal}{\emph{Advances in neural information processing systems}}  \bibinfo{volume}{32} (\bibinfo{year}{2019}).
\newblock


\bibitem[Sohrabizadeh et~al\mbox{.}(2023)]%
        {sohrabizadeh2023robust}
\bibfield{author}{\bibinfo{person}{Atefeh Sohrabizadeh}, \bibinfo{person}{Yunsheng Bai}, \bibinfo{person}{Yizhou Sun}, {and} \bibinfo{person}{Jason Cong}.} \bibinfo{year}{2023}\natexlab{}.
\newblock \showarticletitle{Robust GNN-based representation learning for HLS}. In \bibinfo{booktitle}{\emph{2023 IEEE/ACM International Conference on Computer Aided Design (ICCAD)}}. IEEE, \bibinfo{pages}{1--9}.
\newblock


\bibitem[Wu et~al\mbox{.}(2022a)]%
        {wu2022lostin}
\bibfield{author}{\bibinfo{person}{Nan Wu}, \bibinfo{person}{Jiwon Lee}, \bibinfo{person}{Yuan Xie}, {and} \bibinfo{person}{Cong Hao}.} \bibinfo{year}{2022}\natexlab{a}.
\newblock \showarticletitle{Lostin: Logic optimization via spatio-temporal information with hybrid graph models}. In \bibinfo{booktitle}{\emph{2022 IEEE 33rd International Conference on Application-specific Systems, Architectures and Processors (ASAP)}}. IEEE, \bibinfo{pages}{11--18}.
\newblock


\bibitem[Wu et~al\mbox{.}(2022b)]%
        {wu2022ai}
\bibfield{author}{\bibinfo{person}{Nan Wu}, \bibinfo{person}{Yuan Xie}, {and} \bibinfo{person}{Cong Hao}.} \bibinfo{year}{2022}\natexlab{b}.
\newblock \showarticletitle{AI-assisted Synthesis in Next Generation EDA: Promises, Challenges, and Prospects}. In \bibinfo{booktitle}{\emph{2022 IEEE 40th International Conference on Computer Design (ICCD)}}. IEEE, \bibinfo{pages}{207--214}.
\newblock


\bibitem[Xu et~al\mbox{.}(2019)]%
        {xu2018powerful}
\bibfield{author}{\bibinfo{person}{Keyulu Xu}, \bibinfo{person}{Weihua Hu}, \bibinfo{person}{Jure Leskovec}, {and} \bibinfo{person}{Stefanie Jegelka}.} \bibinfo{year}{2019}\natexlab{}.
\newblock \showarticletitle{How powerful are graph neural networks?}. In \bibinfo{booktitle}{\emph{International Conference on Learning Representations}}.
\newblock


\bibitem[Yang et~al\mbox{.}(2022a)]%
        {yang2022prediction}
\bibfield{author}{\bibinfo{person}{Chenghao Yang}, \bibinfo{person}{Yinshui Xia}, {and} \bibinfo{person}{Zhufei Chu}.} \bibinfo{year}{2022}\natexlab{a}.
\newblock \showarticletitle{The prediction of the quality of results in Logic Synthesis using Transformer and Graph Neural Networks}.
\newblock \bibinfo{journal}{\emph{arXiv preprint arXiv:2207.11437}} (\bibinfo{year}{2022}).
\newblock


\bibitem[Yang et~al\mbox{.}(2022b)]%
        {yang2022versatile}
\bibfield{author}{\bibinfo{person}{Shuwen Yang}, \bibinfo{person}{Zhihao Yang}, \bibinfo{person}{Dong Li}, \bibinfo{person}{Yingxueff Zhang}, \bibinfo{person}{Zhanguang Zhang}, \bibinfo{person}{Guojie Song}, {and} \bibinfo{person}{Jianye Hao}.} \bibinfo{year}{2022}\natexlab{b}.
\newblock \showarticletitle{Versatile multi-stage graph neural network for circuit representation}.
\newblock \bibinfo{journal}{\emph{Advances in Neural Information Processing Systems}}  \bibinfo{volume}{35} (\bibinfo{year}{2022}), \bibinfo{pages}{20313--20324}.
\newblock


\bibitem[Yu(2023)]%
        {yu2023machine}
\bibfield{author}{\bibinfo{person}{Bei Yu}.} \bibinfo{year}{2023}\natexlab{}.
\newblock \showarticletitle{Machine Learning in EDA: When and How}. In \bibinfo{booktitle}{\emph{2023 ACM/IEEE 5th Workshop on Machine Learning for CAD (MLCAD)}}. IEEE, \bibinfo{pages}{1--6}.
\newblock


\end{thebibliography}

\end{document}